\pdfoutput=1

\documentclass[11pt]{article}

\usepackage[final]{acl}

\usepackage{amsmath}
\usepackage{latexsym}

\usepackage[english]{babel}
\usepackage[T2A, T1]{fontenc}
\usepackage{newtxtext,newtxmath}

\usepackage[utf8]{inputenc}

\usepackage{microtype}

\usepackage{inconsolata}

\usepackage{graphicx}

\title{Vuyko Mistral: Adapting LLMs for Low-Resource Dialectal Translation}

\author{
  \textbf{Roman Kyslyi\textsuperscript{1}},
  \textbf{Yuliia Maksymiuk\textsuperscript{2}},
  \textbf{Ihor Pysmennyi\textsuperscript{1}}
\\
\textsuperscript{1}National Technical University of Ukraine "Igor Sikorsky Kyiv Polytechnic Institute" \\
\textsuperscript{2}Ukrainian Catholic University
\\
\small{
  \textbf{Correspondence:} \href{mailto:kyslyi.roman@lll.kpi.ua}{kyslyi.roman@lll.kpi.ua}, 
  \href{mailto:yuliia.maksymyuk@ucu.edu.ua}{yuliia.maksymyuk@ucu.edu.ua}, 
  \href{mailto:pysmennyi.ihor@lll.kpi.ua}{pysmennyi.ihor@lll.kpi.ua}
}
}

\begin{document}
\maketitle
\begin{abstract}
In this paper we introduce the first effort to adapt large language models (LLMs) to the Ukrainian dialect (in our case Hutsul), a low-resource and morphologically complex dialect spoken in the Carpathian Highlands. We created a parallel corpus of 9852 dialect-to-standard Ukrainian sentence pairs and a dictionary of 7320 dialectal word mappings. We also addressed data shortage by proposing an advanced Retrieval-Augmented Generation (RAG) pipeline to generate synthetic parallel translation pairs, expanding the corpus with 52142 examples. We have fine-tuned multiple open-source LLMs using LoRA and evaluated them on a standard-to-dialect translation task, also comparing with few-shot GPT-4o translation. In the absence of human annotators, we adopt a multi-metric evaluation strategy combining BLEU, chrF++, TER, and LLM-based judgment (GPT-4o). The results show that even small(7B) finetuned models outperform zero-shot baselines such as GPT-4o across both automatic and LLM-evaluated metrics. All data, models, and code are publicly released at: \url{https://github.com/woters/vuyko-hutsul}.
\end{abstract}

\section{Introduction}

Despite recent advances in large language models (LLMs), most research and applications remain centered on high-resource languages and their standard variants \cite{li2024language}. This imbalance has significant consequences for linguistic diversity, particularly for underrepresented dialects that lack sufficient textual resources and standardized orthographies\cite{zhong2024opportunities}. Despite being an integral part of the linguistic identity of many communities, dialects are often excluded from NLP tools and research, limiting their accessibility and risking further marginalization and extinction\cite{syed2023quantifying}.

Language technologies and especially LLMs are playing a growing role in the preservation of endangered and underrepresented languages. While much attention has focused on major indigenous languages (e.g., Māori, Quechua, Inuktitut)\cite{trudgill2003dialect, cooper2024attending}, dialects of national languages are often overlooked despite facing similar pressures of attrition and assimilation. Dialectal variants, particularly in post-Soviet contexts, often carry suppressed cultural identities that are not reflected in the standard language. These dialects are not only linguistically rich but also culturally vital and deserve computational attention.

Ukrainian, a language low in resources according to global standards itself\cite{kiulian2024from}, exhibits rich internal variation, with dialects such as Hutsul, Boyko and Lemko\footnote{\url{https://en.wikipedia.org/wiki/Hutsuls}} preserving unique phonetic, lexical and grammatical characteristics. Among these, the Hutsul dialect, spoken in the Carpathian Mountains, is one of the most linguistically distinct and has the most written sources. From the culture standpoint, Hutsul dialect has a great significance as it encapsulates traditions, folklore, and a unique worldview, playing a central role in community identity.

However, the lack of digitized corpora, dictionaries, and processing tools makes it practically invisible to modern LLMs.

Here are some of the linguistic Characteristics of Hutsul dialect:
\begin{itemize}
    \item \textit{Phonetics}: vowel transformations, such as changing vowels {\fontencoding{T2A}\selectfont"є"} instead of 
 {\fontencoding{T2A}\selectfont"а", "я"(ya)} 
 (example:{\fontencoding{T2A}\selectfont"як" → "єк", "ягода" → "єгода"} (``yak'' → ``yek'', ``yahoda'' → ``yehoda'')).
    \item \textit{Morphology}: unique case endings ({\fontencoding{T2A}\selectfont-єдь}, -ci) ('-yed', '-si') and preserved dual forms {\fontencoding{T2A}\selectfontдві яблуціi} (``two apples'', with dual form ``yablutsi'' instead of plural ``yabluka'').
    \item \textit{Lexicon}: Romanian, Polish and German borrowings such as {\fontencoding{T2A}\selectfont\textit{"бринза"}} (cheese) and {\fontencoding{T2A}\selectfont\textit{"шпацірувати"}} (go for a walk).\footnote{\url{https://en.wikipedia.org/wiki/Eastern_Romance_influence_on_Slavic_languages}}
\end{itemize}

\begin{figure}[h]
    \centering
    \includegraphics[width=0.95\linewidth]{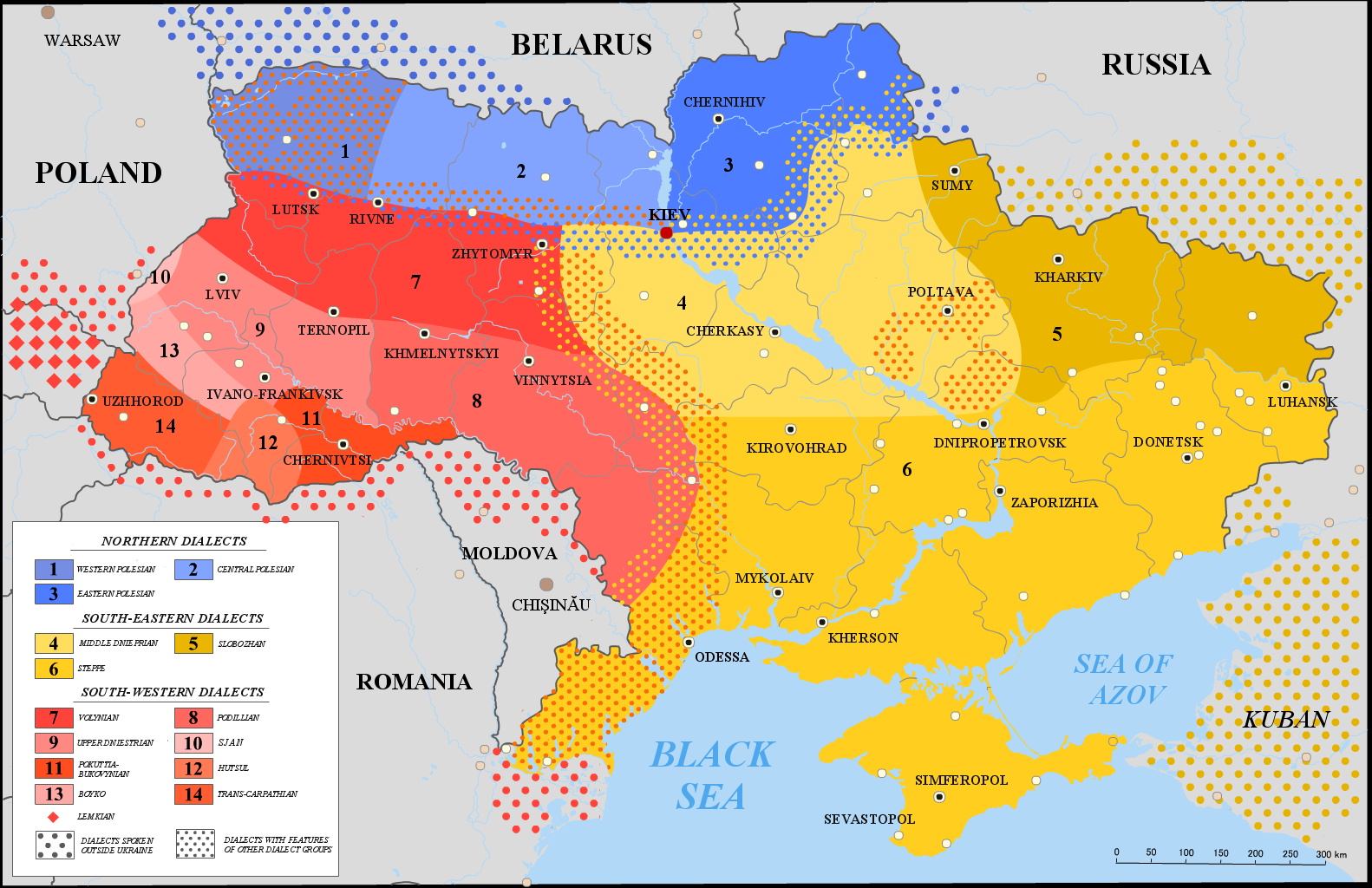}
    \caption{Map of Ukrainian dialects. The Hutsul dialect is located in the southwestern Carpathian region.
    Source: \href{https://en.wikipedia.org/wiki/Ukrainian_dialects\#/media/File:Map_of_Ukrainian_dialects_en.png}{Wikipedia}}
    \label{fig:dialect-map}
\end{figure}

In this work, we present an effort to adapt LLMs to the Hutsul dialect of Ukrainian, addressing both data shortage and modeling challenges. Our contributions are:
\begin{itemize}
    \item A new parallel corpus of original Hutsul-Ukrainian (9852 sentence pairs), dictionary of 7320 dialectal word mappings and also synthetically extended corpus (52142 sentence pairs), using an advanced RAG approach (detailed described below).
    \item Fine-tuning of several open-source LLMs for Ukrainian to Hutsul dialect translation task.
\end{itemize}

We frame our task as standard-to-dialect translation, in which model has to take standard Ukrainian as input and produce grammatically correct (or as close as possible) Hutsul dialect. Our models show that it is feasible to address such translation with limited parallel data and targeted augmentation strategies.

To our knowledge, this is the first work that tries to adapt LLMs to a Ukrainian dialect and among the few globally addressing dialect-to-standard generation using synthetic augmentation.

\section{Related Work}
\subsection{Dialectal NLP and Language Variation}
In recent years we can see growing interest in dialect modeling, particularly for Arabic \citep{zampieri2017arabic}, German \citep{hollenstein2020germans}, and Romance languages \citep{ramponi2021italian}. These efforts mainly focus on classification, generation, and translation between dialects and their standard variants. However, most research remains concentrated on high-resource languages and dialects with pre-existing NLP resources.
At the same time, within Ukrainian language, dialectal NLP remains underexplored. The VarDial workshop series\citep{vardial2024overview} has supported work for different Slavic languages on related tasks such as cross-dialect machine translation and morphological modeling \citep{blokland2024sami,afarli2024norwegian}. For example, \citet{afarli2024norwegian} explore MT between Bokmål and Nynorsk, while \citet{blokland2024sami} tackle dialectal variation in North Sámi. The SIGMORPHON 2023 shared task \citep{kirov2023sigmorphon} highlighted the importance of lexicon-based inflection modeling for low-resource morphological variants.
\subsection{Dialect-to-Standard Normalization}
The task of normalizing dialectal language to its standard form has been explored using various alignment techniques. \citet{scherrer2023character} evaluated character alignment methods for sentence-level standardization of dialect transcriptions across Finnish, Norwegian, and Swiss German. The study compared approaches from dialectometry, speech processing, and machine translation, finding that trained alignment methods offered only small benefits over simple Levenshtein distance. This suggests that simple yet robust statistical methods may still provide strong baselines in resource-constrained dialectal settings. Moreover, the study underlines the need for tailored preprocessing and alignment tools when working with highly variable and phonetically rich dialect data.
\subsection{LLMs and Dialect Adaptation}
Several recent studies investigate adapting LLMs to dialectal data. \citet{held2024tada} propose task-agnostic adapters for dialect adaptation, while \citet{liu2024dada} introduce dynamic adapter aggregation based on linguistic distance. Tokenizer retrofitting for morphologically rich dialects is explored by \citet{csaki2023tokenizer}. These works demonstrate that both architectural and data-centric interventions are necessary for effective adaptation.
However, these approaches are primarily evaluated on English dialects (e.g., African American English, Indian English) using curated corpora such as Multi-VALUE \citep{lin-etal-2021-multivalue}, and rely on annotated dialect-to-standard pairs, which are rarely available for under-resourced dialects.
\subsection{Low-Resource and Synthetic Data Techniques}
Our work also benefits from previous research in low-resource translation and text generation with synthetic data. \citet{gudibande2023synthetic} and \citet{garcia2024hallucinate} propose retrieval-based or prompt-based augmentation techniques to bootstrap performance in limited-data settings. At the same time we propose our own approach for generating synthetic data using advanced RAG techniques.

\section{Dataset Creation}

\subsection{Parallel Corpus Collection}

We constructed the first parallel corpus for the Hutsul dialect and standard Ukrainian by combining multiple sources and annotation strategies. The dataset includes 9852 sentence pairs, manually aligned at the sentence level. Source texts in Hutsul were collected from publicly available books, ethnographic transcripts, folklore websites, and dialect blogs. A significant portion of the dataset is based on the novel  {\fontencoding{T2A}\selectfont\textit{"Дідо Иванчік"}}(Dido Yvanchik) by Petro Shekeryk-Donykiv\footnote{\url{https://pl.wikipedia.org/wiki/Petro_Szekeryk-Donykiw}}, a foundational literary work written in authentic Hutsul. 
We are especially grateful to the publishing house {\fontencoding{T2A}\selectfont\textit{Дискурс}} and translator {\fontencoding{T2A}\selectfont\textbf{Іван Андрусяк}}, who kindly approved the use of their modern standard Ukrainian translation for academic purposes.

Standard Ukrainian references in the dataset were either manually translated or sourced from bilingual editions where available. To ensure linguistic diversity, we tried to included examples from both everyday conversation and stylized narrative texts (e.g., folk tales, songs, etc.), but due to data shortage some topics remain uncovered.

\subsection{Lexical Resource}

We compiled a Hutsul‑to‑Ukrainian dictionary that now contains about 7 300 word pairs. The work started from the vocabulary that appears in the book {\fontencoding{T2A}\selectfont\textit{"Дідо Иванчік"}}(Dido Yvanchik), but we soon enlarged it with data taken from websites that explain Hutsul dialect words.
Among the most useful web sources were:

\begin{itemize}
\item "Dictionary of Hutsul Words"\footnote{\url{https://karnauhova.at.ua/publ/1-1-0-3}}.
\item "Hutsul Hovir"\footnote{\url{https://rakhiv-mr.gov.ua/hutsulskyj-hovir/}}.
\item "Dictionary of Ukrainian Dialects of the Carpathian Region"\footnote{\url{https://evrika.if.ua/88/}}.
\item "Explanatory Dictionary of Hutsul Dialects" by Petro Havuka\footnote{\url{https://evrika.if.ua/1565/}}.
\item "Hutsul dictionary". \footnote{\url{http://www.webteka.com/hutsul-language/}}
\end{itemize}

All these pages were automatically scraped.
The raw text contained a lot of noise: strange characters, extra commentary, uneven tabulation, and inconsistent separators between the Hutsul entry and and its Ukrainian translation. We wrote simple cleaning scripts, converted everything to a single CSV file, and then manually checked the list to remove the last errors. The final result is a clean lexicon with 7 320 Hutsul–Ukrainian pairs. Each entry includes standard and dialectal word forms. 

Despite this effort, the lexicon remains biased toward the vocabulary found in literature and folkloric domains. Due to the shortage of Hutsul texts on topics like news, science, or politics, our dataset lacks sufficient lexical diversity in those domains.

\subsection{Synthetic Data via Advanced RAG}

To overcome shortage of written sources in Hutsul dialect, we developed an advanced RAG pipeline to generate additional Hutsul-standard sentence pairs. 
The foundation of this pipeline was the dialectal novel {\fontencoding{T2A}\selectfont\textit{"Дідо Иванчік"}}(Dido Yvanchik), which served as both the primary corpus for retrieval and the source of linguistic examples. We used GPT-4o to build a RAG module capable of retrieving semantically related Hutsul sentences. For each generation step, a prompt was created containing linguistic transformation rules representative of Hutsul phonological and lexical variation. 

The construction of the RAG pipeline involved several steps:

\begin{enumerate}
    \item \textbf{Grammar Rule Extraction:} Using {\fontencoding{T2A}\selectfont\textit{"Дідо Иванчік"}}(Dido Yvanchik) as input, we prompted GPT-4o to extract and structure grammatical transformation rules characteristic of the Hutsul dialect. These included phonological shifts, morphological alternations, and syntactic reordering. We augmented these rules with material from Wikipedia and Models of Word Formation in Hutsul Dialects \citet{greshchuk2016models} to create a comprehensive prompt template (see Figure 2).
    
    \item \textbf{Indexing via RAG:} We indexed the {\fontencoding{T2A}\selectfont\textit{"Дідо Иванчік"}}(Dido Yvanchik) corpus into a retrieval system to serve as a reference base for generating dialectal outputs using text-embedding-3-large\footnote{\url{https://platform.openai.com/docs/models/text-embedding-3-large}}.
    
    \item \textbf{Candidate Sentence Selection:} Standard Ukrainian sentences were sampled from the UberText corpus (\citet{chaplynskyi-2023-introducing}). For each such sentence, we used the RAG module to retrieve the top-3 semantically similar Hutsul-like sentences from {\fontencoding{T2A}\selectfont\textit{"Дідо Иванчік"}}(Dido Yvanchik).
    
    \item \textbf{Prompt Construction:} The retrieved examples were inserted into the prompt template along with the standard Ukrainian sentence as the source for translation.
    
    \item \textbf{Dialect Generation:} GPT-4o was instructed to produce a Hutsul translation of the input sentence using the provided grammar rules and examples as context (see Figure 3).
\end{enumerate}

Below is a main part or our rule-based prompt (Full prompt can be found here: \url{https://github.com/woters/vuyko-hutsul/blob/main/prompts/hutsul_rules_prompt.txt}):

\begin{quote}
\small
\texttt{
Here are Grammatical Rules for Converting Ukrainian Text into the Hutsul Dialect:\\
1. Vowel Shifts:\\
- {\fontencoding{T2A}\selectfont``як'' → ``єк''} \textit{(``yak'' → ``yek'')}\\
- {\fontencoding{T2A}\selectfont``яблуко'' → ``єблуко''} (``yabluko'' → ``yebluko'')\\
- {\fontencoding{T2A}\selectfont``йдеш'' → ``єдеш''} (``yidesh'' → ``yedesh'')\\[0.3em]
2. Consonant Transformations:\\
- {\fontencoding{T2A}\selectfont``дівка'' → ``ґівка''} (``divka'' → ``givka'')\\
- {\fontencoding{T2A}\selectfont``чого'' → ``чьо''} (``choho'' → ``cho'')\\
- {\fontencoding{T2A}\selectfont``ти'' → ``ци''} (``ty'' → ``tsy'')\\\\[0.3em]
3. Word Order and Syntax:\\
- {\fontencoding{T2A}\selectfont``Я тебе люблю'' → ``Люблю я тебе''} (``I love you'' → ``Love I you'')\\
- {\fontencoding{T2A}\selectfont``Він сміється'' → ``Він смієтси''} (``He is laughing'' → ``He laugh-reflexive'')\\
- {\fontencoding{T2A}\selectfont``Ти знаєш?'' → ``Ци ти знаєш?''} (``Do you know?'' → ``Do you know?'' with dialectal marker ``tsy'')\\[0.3em]
Apply only contextually appropriate transformations. 
}
\end{quote}

This process have created some data alignment challenges in the generated dataset. To address these challenges and also to clean generated dataset we have developed a hybrid alignment strategy. First we leveraged the expected textual similarity between a language and its dialect using difflib's SequenceMatcher\footnote{\url{https://docs.python.org/3/library/difflib.html}}. This approach directly compares character sequences, effectively identifying pairs even with minor dialectal variations. Pairs falling below a similarity threshold of 0.45 was removed from the dataset. To measure quality of remained sentence pairs we have
used several statistical metrics as described by \citet{scherrer2023character}:

\begin{itemize}
    \item \textbf{U-src} – proportion of unaligned source characters,
    \item \textbf{U-tgt} – proportion of unaligned target characters,
    \item \textbf{X} - proportion of crossing alignment pairs (swaps)
\end{itemize}

These metrics were calculated over symmetrized alignment pairs obtained with fast align\cite{dyer-etal-2013-simple}. We have compared alignment metrics across three datasets: the original manually annotated dataset (mainly from {\fontencoding{T2A}\selectfont\textit{"Дідо Иванчік"}}(Dido Yvanchik)), raw synthetically generated dataset, and the filtered synthetic dataset. 

Before filtering, the synthetic data already exhibited lower proportions of unaligned source and target words (U-src=0.139, U-tgt=0.136) compared to the original data (U-src=0.260, U-tgt=0.265). However, it presented a higher proportion of crossing alignments (X=0.033 vs. 0.022 original), indicating increased structural variability.

To improve the quality of our generated dataset, we applied alignment-based filtering - for each sentence pair, we have used previously calculated statistics(U-src, U-tgt and X) and we empirically defined a thresholds for them: $U\text{-src} < 0.1$, $U\text{-tgt} < 0.1$, and $X < 0.2$.

\begin{table}[ht]
\centering
\small
\begin{tabular}{lccc}
\hline
\textbf{Metric} & \textbf{Original} & \textbf{Synthetic} & \textbf{Synthetic} \\
 & \textbf{Dataset} & \textbf{(Raw)} & \textbf{(Filtered)} \\
\hline
U-src & 0.260 & 0.139 & \textbf{0.005} \\
U-tgt & 0.265 & 0.136 & \textbf{0.005} \\
X & 0.022 & 0.033 & \textbf{0.019} \\
\hline
\end{tabular}
\caption{Alignment quality metrics comparison between the original dataset, raw synthetic dataset, and synthetic dataset after alignment-based filtering.}
\label{tab:alignment-metrics-comparison}
\end{table}

Any sentence pair that exceeded one or more of these thresholds was excluded from the final data set. This procedure removed inconsistent examples, reducing the number of reorderings, and improving alignment. As the result we got a better quality synthetic dataset with better structural alignment, as demonstrated by the comparative metrics in Table~\ref{tab:alignment-metrics-comparison}.

Although we acknowledge that the obtained synthetic data has some variation and lack of certain lexical phrases present in authentic dialectal speech, its inclusion is justified by shortage of Hutsul textual resources. This filtering step effectively improved the consistency and reliability of the synthetic dataset and added additional 52142 phrase pairs to our training dataset. 

\begin{figure}[h]
    \centering
    \includegraphics[width=0.9\linewidth]{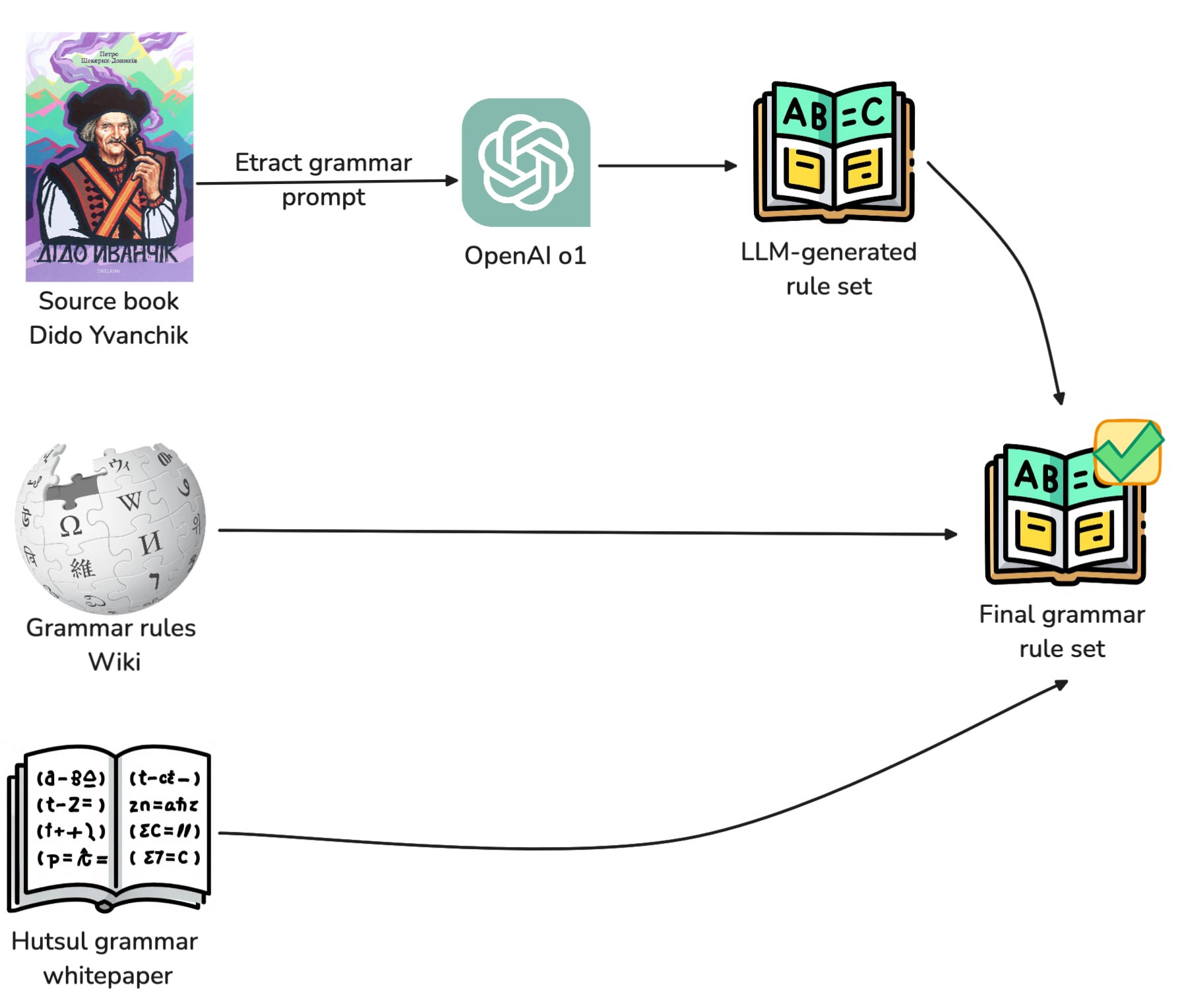} 
    \caption{Overview of the rules generation pipeline based on {\fontencoding{T2A}\selectfont\textit{"Дідо Иванчік"}}(Dido Yvanchik), Wikipedia, and \citet{greshchuk2016models}.}
    \label{fig:rules}
\end{figure}

Although this approach enabled us to significantly enlarge the dataset, it also introduced certain limitations. Specifically, the synthetic data reflects the lexical and topical range of the source corpus, which lacks modern domains such as aviation, technology, news and politics. 

As a result, lexical coverage in these areas remains quite sparse or absent (even after generation, words still remain the same as they are in standard Ukrainian). To avoid introducing hallucinated vocabulary, we deliberately excluded modern news and web-based corpora from the generation process.

\begin{figure}[h]
    \centering
    \includegraphics[width=1\linewidth]{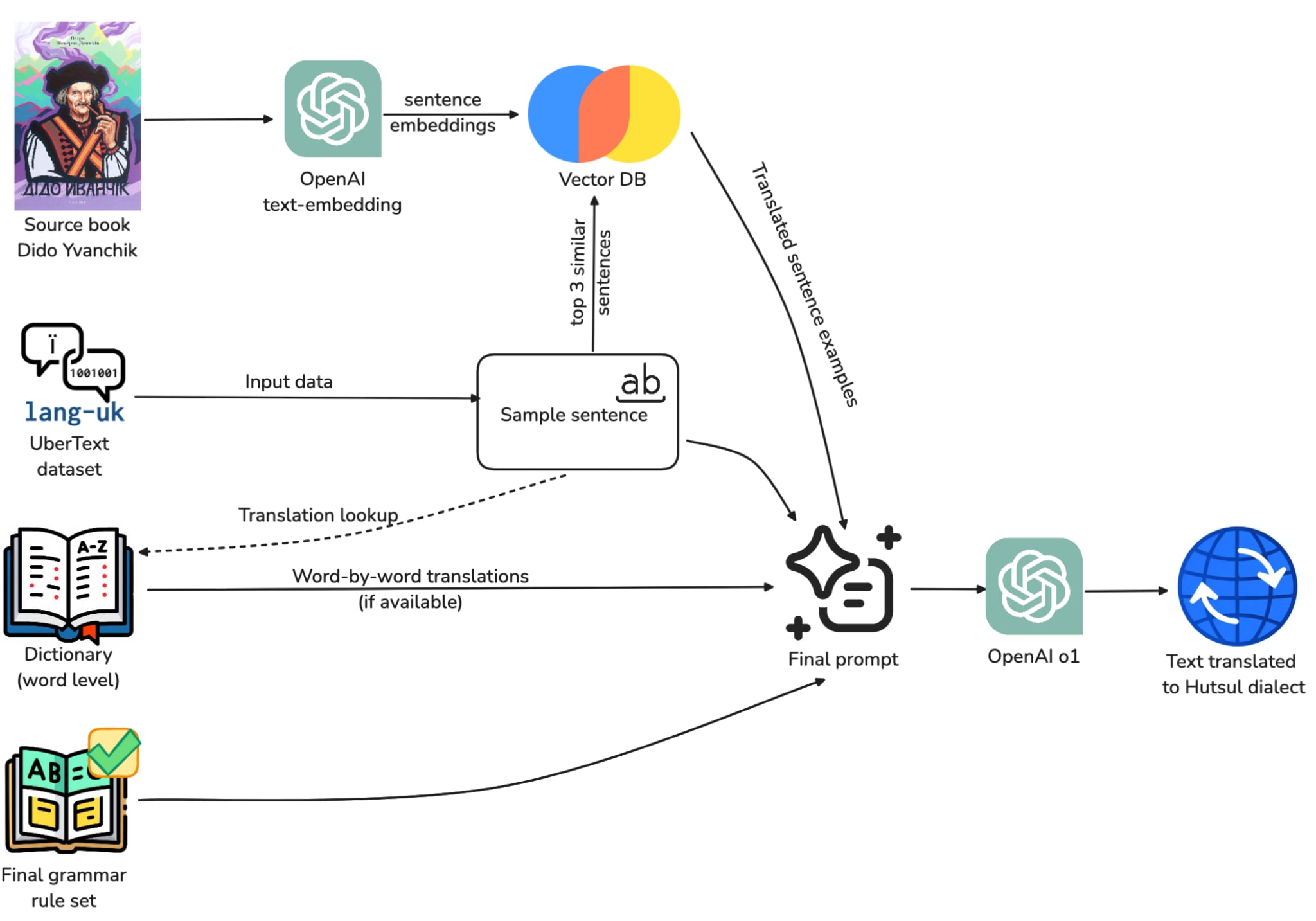} 
    \caption{Overview of the synthetic data generation pipeline: A RAG system using {\fontencoding{T2A}\selectfont\textit{"Дідо Иванчік"}}(Dido Yvanchik) and UberCorpus retrieves and prompts GPT-4o to generate high-quality Hutsul-Ukrainian pairs.}
\end{figure}

\subsection{Data Splits and Availability}

The final corpus was split into 80\% training, 10\% validation, and 10\% test sets. Test and validation sets contain only human-annotated sentence pairs from {\fontencoding{T2A}\selectfont\textit{"Дідо Иванчік"}}(Dido Yvanchik).

\section{Fine-Tuning}

To adapt large language models (LLMs) to the  Ukrainian-to-Hutsul translation task, we used parameter-efficient fine-tuning using LoRA \citep{hu2022lora}. 

We fine-tuned two state-of-the-art open-source models in the 7B–13B parameter range (as we considered our training resources and that the model should be not too big to be able to run locally):

\begin{itemize}
    \item \textbf{Mistral-7B-Instruct v0.3}\footnote{\url{https://huggingface.co/mistralai/Mistral-7B-Instruct-v0.3}} – Chosen for its performance-to-size ratio. It outperforms some larger models on many benchmarks, supports multilingual instructions, and includes explicit support for Ukrainian \citep{mistral2023mistral7b}.
    
    \item \textbf{LLaMA-3.1 8B Instruct}\footnote{\url{https://huggingface.co/meta-llama/Llama-3.1-8B-Instruct}} – The instruction-tuned version of LLaMA 3.1 8B. This model has a strong multilingual support and improved instruction-following ability, making it a good candidate for low-resource translation \citep{touvron2024llama3}.
\end{itemize}

Models were selected based on the following criteria:
\begin{itemize}
    \item Tokenizer support – Both models use tokenizers with fallback strategies for rare or out-of-vocabulary tokens, enabling good handling of Cyrillic-based dialects.
    \item Multilingual capabilities – Mistral-7B-Instruct v0.3 explicitly lists Ukrainian among supported languages. LLaMA-3.1 8B Instruct has shown strong generalization capabilities\footnote{\url{https://huggingface.co/blog/akjindal53244/llama31-storm8b}}.
    \item Open licensing and reproducibility – Both models are publicly available under open-source licenses.
    \item Feasibility on a single GPU – Using LoRA or QLoRA, all selected models can be fine-tuned within the a single not very big GPU.
\end{itemize}

We also considered other multilingual models such as BLOOMZ (7.1B)\footnote{\url{https://huggingface.co/bigscience/bloomz-7b1}}  and NLLB-200 (3.3B)\footnote{\url{https://huggingface.co/facebook/nllb-200-3.3B}}, which offer extensive language coverage. However, these models either underperformed on general language modeling tasks or lacked strong generation quality compared to selected models. Recent benchmarks demonstrate that Mistral-7B-Instruct-v0.3\footnote{\url{https://huggingface.co/mistralai/Mistral-7B-Instruct-v0.3}} matches or surpasses larger models in translation tasks, particularly in low-resource and instruction-tuned settings \citep{wu2023benchmarking}.

\subsection{Fine-Tuning Setup}

Each model was trained for 3 epochs using LoRa on two dataset variants (complete setup can be found in the Guthub\footnote{\url{https://github.com/woters/vuyko-hutsul}}
): (1) a manually created Hutsul–Ukrainian parallel corpus, and (2) an extended version that included combined manual and filtered synthetic data. 

\section{Evaluation}

\subsection{Metrics}

Evaluating dialectal machine translation is not a simple task, as standard reference-based metrics may penalize correct lexical variation. To insure translation quality we calculated the following widely used metrics:

\begin{itemize}
    \item \textbf{BLEU} \citep{papineni2002bleu} - a precision-based metric measuring n-gram overlap between hypothesis and reference. While widely used, it may penalize valid lexical and syntactic variations common in dialects.
    
    \item \textbf{chrF++} \citep{popovic2015chrf} computes character n-gram F-scores and has been shown to outperform BLEU on morphologically rich and non-standard languages. It is more robust to minor spelling or inflectional differences, making it particularly suitable for dialectal text.
    
    \item \textbf{TER} (Translation Edit Rate) \citep{snover2006study} quantifies the number of edits required to convert the system output into the reference. It captures structural divergence and penalizes reordering errors.
\end{itemize}

Each metric emphasizes different aspects of translation quality:
\begin{itemize}
    \item \textit{BLEU} reflects n-gram precision,
    \item \textit{chrF++} captures morphological similarity and recall,
    \item \textit{TER} penalizes structural mismatches
\end{itemize}

We apply these metrics to test set of manually translated Ukrainian–Hutsul sentence pairs (1900 pairs). Rather than aggregating them into a single score, we interpret them jointly to understand different behavioral aspects of each model. For instance, a high chrF++ score alongside a low BLEU score may indicate valid variation in surface realization.

As mentioned before, while these metrics provide a useful baseline, they often struggle to evaluate dialectal outputs. So, following the framework of \citet{aepli2023benchmark}, we incorporate LLMs as evaluators.

We prompt GPT-4o model to rate model outputs along three axes:

\begin{itemize}
    \item \textbf{Fluency}: grammaticality and naturalness in the Hutsul dialect.
    \item \textbf{Adequacy}: preservation of the source sentence’s meaning.
    \item \textbf{Dialectal Quality}: consistency with known lexical, phonological, and morphosyntactic properties of Hutsul.
\end{itemize}

Each evaluation is performed in a zero-shot setting. We thought about including some grammatical rules into the prompt, but to avoid creation of potential bias through this rules decided to use zero-shot instead. 

LLM receives the source, model output, and a reference translation and returns scores from 1 (poor) to 5 (excellent). The prompt is structured as follows:

\begin{quote}
You are a linguistic expert evaluating machine-translated dialectal text. Rate the translation on the following dimensions:

1. Fluency (1–5): Is the output grammatically correct and natural in the target dialect?

2. Adequacy (1–5): Does the output preserve the meaning of the original source?

3. Dialectal Quality (1–5): Does the output reflect the expected phonological, lexical, and grammatical properties of the Hutsul dialect?

Return your answer in this exact JSON format:

\{ "fluency": x, "adequacy": y, "dialect": z \}

Do not explain your ratings.

Source (Standard Ukrainian): <source sentence>

Model Output (Hutsul): <model prediction>

Reference (Hutsul): <reference sentence>
\end{quote}

As we didn't have an opportunity to perform a human evaluation for our translation, and considering that standard reference-based metrics may not be a good fit for the dialect translation\citep{aepli2023benchmark}, we have used the LLM-based adequacy and dialect scores as our primary evaluation metrics. These are better aligned with human intuition and more tolerant of different variations than BLEU or TER.

Automatic metrics are used in a supporting role to identify trends such as reordering or character-level similarity. We report all metrics side-by-side. This multi-metric approach enables a more holistic interpretation of model behavior, especially in the absence of human raters.

\subsection{Baselines}

We compared our fine-tuned models against the GPT-4o baseline. Queried via the OpenAI API, prompted to translate standard Ukrainian into the Hutsul dialect. To ensure consistency and lexical coverage, we used the same RAG context and dictionary entries as in our synthetic generation pipeline.

We did not include non fine-tuned Mistral or LLaMA models as baselines, since their performance in dialect generation tasks was much worse. Due to their small size, their instruct tuning is insufficient for zero-shot generation in underrepresented languages or dialects.

\subsection{Results}

As mentioned earlier, we evaluate our models using both automatic metrics and LLM-based judgments. Table~\ref{tab:results} presents the BLEU, chrF++, TER scores and GPT-4o as an LLM-based judge, rating each output on a 1–5 scale for fluency, adequacy, and dialectal quality scores computed on a held-out test set of 1900 sentences.

\begin{table*}[ht]
\centering
\begin{tabular}{lcccccc}
\hline
\textbf{Model} & \textbf{BLEU} & \textbf{chrF++} & \textbf{TER} & \textbf{Fluency} & \textbf{Adequacy} & \textbf{Dialect} \\
\hline
GPT-4o & 56.64 & 65.90 & 34.34 & 3.76 & 4.30 & 3.22 \\
LLaMA (manual annotated + synthetic) & 69.02 & 74.92 & \textbf{22.90} & 4.11 & 4.72 & 3.33 \\
LLaMA (manual annotated only) & 59.98 & 72.61 & 28.62 & 4.13 & 4.72 & 3.38 \\
Mistral (manual annotated only) & 62.36 & 75.65 & 28.62 & 4.14 & \textbf{4.74} & 3.35 \\
Mistral (manual annotated + synthetic) & \textbf{74.35} & \textbf{81.89} & \textbf{22.90} & \textbf{4.18} & \textbf{4.72} & \textbf{3.60} \\
\hline
\end{tabular}
\caption{Automatic and LLM-based evaluation results. BLEU, chrF++, and TER are computed with \texttt{sacreBLEU}. Fluency, adequacy, and dialect quality are rated by GPT-4o (1–5 scale).}
\label{tab:results}
\end{table*}

From the results we can see that all fine-tuned models outperform the GPT-4o baseline for every metric. Mistral fine-tuned on combined manually collected and synthetic data performs best overall, with the highest BLEU (74.35), chrF++ (81.89), and dialect rating (3.60). While adequacy scores remain stable across all models ($\approx$4.7), dialectal accuracy varies more substantially and proves most sensitive to the source of training data. Also we can see that both, LLaMA and Mistral trained on combined synthetic and manually annotated data show strong scores on automatic metrics but slightly underperform on dialectal quality, highlighting the limitations of our method of generating synthetic data.

\vspace{1em}

\subsection{Qualitative Examples}
Below we show an example depicting LLM-calculated scores over real data along with respective BLEU, chrF++, and TER metrics. This demonstrates that even small fine-tuned models are slightly better at preserving dialect-specific meaning and lexicon than zero-shot commercial models, but still far from perfect.
\begin{description}

  \item[Reference (Hutsul):] 
  {\fontencoding{T2A}\selectfont\textit{"Прошумавси у вечєр, єк зробивси в діда в обох хатах гармідер."}}  
  (Eng: \textit{He came to his senses in the evening, after a mess started in both of the grandfather's houses.})

  \item[GPT-4o:] 
  {\fontencoding{T2A}\selectfont\textit{"Отєтавсі аж увечєру, єк зчинивсі в ґіда в обидвох осєдочьках гармидер."}}
  (Eng: \textit{He snapped out of it only in the evening, after a chaos started in both of the old man’s huts.})  
  \\ BLEU: 7.77, chrF++: 32.8, TER: 72.73, Fluency: 4, Adequacy: 4, Dialect: 3

  \item[Mistral (manual + synthetic):]  
  {\fontencoding{T2A}\selectfont\textit{"Отямивси аж увечері, коли зчинився в діда в обох хатах гармидер."}}  
  (Eng: \textit{He came to his senses in the evening, when a chaos broke out in both grandfather's houses.})  
  \\ BLEU: 31.46, chrF++: 46.62, TER: 54.55, Fluency: 4, Adequacy: 5, Dialect: 3

  \item[Mistral (manual only):]  
  {\fontencoding{T2A}\selectfont\textit{"Отємивси аж увечєрі, єк зчєнивси в діда в обох хатах гармидер."}}  
  (Eng: \textit{He came to his senses only in the evening, after stirring up a mess in both of the grandfather’s houses.})  
  \\ BLEU: 34.39, chrF++: 56.92, TER: 45.45, Fluency: 4, Adequacy: 5, Dialect: 4

  \item[LLAMA (manual + synthetic):]  
  {\fontencoding{T2A}\selectfont\textit{"Отямивси аж у вечєрє, коли зчинивси в діда в обох хатях гармидер."}}  
  (Eng: \textit{He came to his senses in the evening, when a mess started in both grandfather’s houses.})  
  \\ BLEU: 20.45, chrF++: 46.42, TER: 63.64, Fluency: 4, Adequacy: 4, Dialect: 3

  \item[LLAMA (manual only):]  
  {\fontencoding{T2A}\selectfont\textit{"Отємивси аж увечері, єк зчинивси у діда в обох хатах гармидер."}} 
  (Eng: \textit{He came to his senses only in the evening, when a mess in the grandfather’s both houses started.})  
  \\ BLEU: 24.71, chrF++: 49.59, TER: 54.55, Fluency: 4, Adequacy: 5, Dialect: 3

\end{description}

\section*{Limitations}

Our work makes first step in Ukrainian dialect adaptation for LLMs, a lot of limitations remain open.

An important limitation is that although we introduced a synthetic data generation pipeline to mitigate limited data availability problem, synthetic translations may lack native fluency or have stylistic inconsistencies, especially for underrepresented topics. This is particularly can be seen in domains not covered by the original corpus, such as politics, technology, etc. where Hutsul lexicon is either very limited or absent. Despite filtering low-quality generations, automatic evaluation metrics still may overestimate linguistic validity.

In addition, evaluation remains challenging. Automatic metrics such as BLEU and chrF++ often penalize valid dialectal variation \citep{garcia2024hallucinate, held2024tada}. To better capture stylistic and synthetic diversity, we use GPT-4o as an LLM-based judge following recent work on LLM-based evaluation frameworks \citep{wang2023llmeval, liu2023gpteval}. However, we note that GPT-4o is not explicitly fine-tuned for dialectal assessment, and its preferences may still align with standard Ukrainian and human evaluation would provide much more reliable assessments.

Also we need to mention that our current methods are tailored to Hutsul, a relatively well-documented dialect within the Ukrainian language. Extension to other dialects or usage of the same approach for other low-resource languages will require adaptation of both the data pipeline and prompting strategies.

\section*{Acknowledgments}
We express our sincere gratitude to {\fontencoding{T2A}\textbf{Іван Андрусяк}}(Ivan Andrusiak) for providing access to his Ukrainian translation of {\fontencoding{T2A}\textit{"Дідо Иванчік"}}(Dido Yvanchik), which served as a cornerstone of our dataset. We also thank the publishing house {\fontencoding{T2A}\textbf{"Дискурс"}}(Dyskurs) and its director {\fontencoding{T2A}\textbf{Василь Карп’юк}}(Vasyl Karpiuk) for their kind permission to use the text and for their continued support of linguistic and cultural preservation initiatives. Their generosity made this research possible.

\bibliography{custom}

\begin{thebibliography}{33}
\providecommand{\natexlab}[1]{#1}

\bibitem[{Aepli et~al.(2023)Aepli, Ebling, and Sennrich}]{aepli2023benchmark}
No{\"e}mi Aepli, Sarah Ebling, and Rico Sennrich. 2023.
\newblock A benchmark for evaluating machine translation metrics on dialects without standard orthography.
\newblock \emph{arXiv preprint arXiv:2311.16865}.

\bibitem[{AI(2023)}]{mistral2023mistral7b}
Mistral AI. 2023.
\newblock Introducing mistral 7b and mixtral.
\newblock \url{https://mistral.ai/news/mistral-7b/}.

\bibitem[{Blokland et~al.(2024)Blokland, Trosterud, and Rueter}]{blokland2024sami}
Rogier Blokland, Trond Trosterud, and Jack Rueter. 2024.
\newblock \href {https://aclanthology.org/2024.vardial-1.11} {Morphological variants in north s{'a}mi dialects}.
\newblock In \emph{Proceedings of the Tenth Workshop on NLP for Similar Languages, Varieties and Dialects (VarDial)}.

\bibitem[{Chaplynskyi(2023)}]{chaplynskyi-2023-introducing}
Dmytro Chaplynskyi. 2023.
\newblock \href {https://aclanthology.org/2023.unlp-1.1} {Introducing {U}ber{T}ext 2.0: A corpus of modern {U}krainian at scale}.
\newblock In \emph{Proceedings of the Second Ukrainian Natural Language Processing Workshop}, pages 1--10, Dubrovnik, Croatia. Association for Computational Linguistics.

\bibitem[{Cooper et~al.(2024)Cooper, Heldreth, and Hutchinson}]{cooper2024attending}
Ned Cooper, Courtney Heldreth, and Ben Hutchinson. 2024.
\newblock \href {https://aclanthology.org/2024.eacl-short.19} {“it's how you do things that matters”: Attending to process to better serve indigenous communities with language technologies}.
\newblock In \emph{Proceedings of the 18th Conference of the European Chapter of the Association for Computational Linguistics (EACL 2024)}, pages 204--211.

\bibitem[{Cs{'a}ki et~al.(2023)}]{csaki2023tokenizer}
G{'a}bor Cs{'a}ki and 1 others. 2023.
\newblock \href {https://aclanthology.org/2024.acl-long.777} {Tokenizer retrofitting for morphologically rich languages}.
\newblock In \emph{Findings of the Association for Computational Linguistics: ACL 2023}.

\bibitem[{Dyer et~al.(2013)Dyer, Chahuneau, and Smith}]{dyer-etal-2013-simple}
Chris Dyer, Victor Chahuneau, and Noah~A. Smith. 2013.
\newblock \href {https://aclanthology.org/N13-1073/} {A simple, fast, and effective reparameterization of {IBM} model 2}.
\newblock In \emph{Proceedings of the 2013 Conference of the North {A}merican Chapter of the Association for Computational Linguistics: Human Language Technologies}, pages 644--648, Atlanta, Georgia. Association for Computational Linguistics.

\bibitem[{Garcia et~al.(2024)}]{garcia2024hallucinate}
Xavier Garcia and 1 others. 2024.
\newblock \href {https://arxiv.org/abs/2402.07836} {Don't hallucinate, retrieve! a survey on retrieval-augmented text generation}.
\newblock \emph{arXiv preprint arXiv:2402.07836}.

\bibitem[{Greshchuk(2016)}]{greshchuk2016models}
Vasyl Greshchuk. 2016.
\newblock \href {https://jgrst.donnu.edu.ua/article/view/3030} {Models of word formation in hutsul dialects (based on the dictionary “hutsul dialectal vocabulary in ukrainian belletristic language”)}.
\newblock \emph{Gramatychni Studii}, 6:272--286.

\bibitem[{Gudibande et~al.(2023)}]{gudibande2023synthetic}
Aditya Gudibande and 1 others. 2023.
\newblock \href {https://arxiv.org/abs/2305.11864} {Synthetic data scaling for low-resource nlp}.
\newblock \emph{arXiv preprint arXiv:2305.11864}.

\bibitem[{Held and Klakow(2024)}]{held2024tada}
Wolfgang Held and Dietrich Klakow. 2024.
\newblock \href {https://aclanthology.org/2024.kallm-1.7} {Tada: Task-agnostic dialect adapters for multilingual transformers}.
\newblock In \emph{Proceedings of the First Workshop on Domain Adaptation for NLP}.

\bibitem[{Hollenstein et~al.(2020)Hollenstein, Fairon, and Snyers}]{hollenstein2020germans}
Nora Hollenstein, C{'e}drick Fairon, and Julie Snyers. 2020.
\newblock \href {https://aclanthology.org/2020.lrec-1.353} {German’s many voices: A corpus of regional variation in german}.
\newblock In \emph{Proceedings of the 12th Language Resources and Evaluation Conference}.

\bibitem[{Hu et~al.(2021)}]{hu2022lora}
Edward~J Hu and 1 others. 2021.
\newblock \href {https://arxiv.org/abs/2106.09685} {Lora: Low-rank adaptation of large language models}.
\newblock \emph{arXiv preprint arXiv:2106.09685}.

\bibitem[{Kinn and {\AA}farli(2024)}]{afarli2024norwegian}
Torodd Kinn and Tor~A. {\AA}farli. 2024.
\newblock \href {https://aclanthology.org/2024.vardial-1.19} {Exploring parallel machine translation for norwegian nynorsk and bokmål}.
\newblock In \emph{Proceedings of the Tenth Workshop on NLP for Similar Languages, Varieties and Dialects (VarDial)}.

\bibitem[{Kirov et~al.(2023)Kirov, Cotterell et~al.}]{kirov2023sigmorphon}
Christo Kirov, Ryan Cotterell, and 1 others. 2023.
\newblock \href {https://aclanthology.org/2023.sigmorphon-1.12} {Sigmorphon 2023 shared task: Morphological inflection in context}.
\newblock In \emph{Proceedings of the 20th SIGMORPHON Workshop}.

\bibitem[{Kiulian et~al.(2024)Kiulian, Polishko, Khandoga, Chubych, Connor, Ravishankar, and Shirawalmath}]{kiulian2024from}
Artur Kiulian, Anton Polishko, Mykola Khandoga, Oryna Chubych, Jack Connor, Raghav Ravishankar, and Adarsh~Arunkumar Shirawalmath. 2024.
\newblock \href {https://arxiv.org/abs/2404.09138} {From bytes to borsch: Fine-tuning gemma and mistral for the ukrainian language representation}.
\newblock \emph{arXiv preprint arXiv:2404.09138}.

\bibitem[{Li et~al.(2024)Li, Shi, Liu, Yang, Liu, and Du}]{li2024language}
Zihao Li, Yucheng Shi, Zirui Liu, Fan Yang, Ninghao Liu, and Mengnan Du. 2024.
\newblock \href {https://arxiv.org/abs/2404.11553} {Language ranker: A metric for quantifying llm performance across high and low-resource languages}.
\newblock \emph{arXiv preprint arXiv:2404.11553}.

\bibitem[{Lin et~al.(2021)Lin, Tetreault et~al.}]{lin-etal-2021-multivalue}
Zi~Lin, Joel Tetreault, and 1 others. 2021.
\newblock \href {https://aclanthology.org/2021.emnlp-main.621} {Multi-value: A multilingual, multi-dialect, and multi-task benchmark for language understanding}.
\newblock In \emph{Proceedings of EMNLP 2021}.

\bibitem[{Liu et~al.(2024)}]{liu2024dada}
Xiang Liu and 1 others. 2024.
\newblock \href {https://arxiv.org/abs/2409.11404} {Dada: Dynamic adapter aggregation for dialectal adaptation}.
\newblock \emph{arXiv preprint arXiv:2409.11404}.

\bibitem[{Liu(2023)}]{liu2023gpteval}
Ziwei et~al. Liu. 2023.
\newblock Gpt-4 as an automatic grader: An evaluation of zero-shot and few-shot prompting for text scoring tasks.
\newblock \emph{arXiv preprint arXiv:2304.02329}.

\bibitem[{Papineni et~al.(2002)Papineni, Roukos, Ward, and Zhu}]{papineni2002bleu}
Kishore Papineni, Salim Roukos, Todd Ward, and Wei-Jing Zhu. 2002.
\newblock Bleu: a method for automatic evaluation of machine translation.
\newblock In \emph{Proceedings of the 40th Annual Meeting of the Association for Computational Linguistics}, pages 311--318.

\bibitem[{Popović(2015)}]{popovic2015chrf}
Maja Popović. 2015.
\newblock chrf: character n-gram f-score for automatic mt evaluation.
\newblock In \emph{Proceedings of the Tenth Workshop on Statistical Machine Translation}, pages 392--395.

\bibitem[{Ramponi and Plank(2021)}]{ramponi2021italian}
Alan Ramponi and Barbara Plank. 2021.
\newblock \href {https://aclanthology.org/2021.eacl-main.211} {Neural multi-dialect language models for zero-shot cross-dialect transfer}.
\newblock In \emph{Proceedings of the 16th Conference of the European Chapter of the Association for Computational Linguistics: Main Volume}.

\bibitem[{Scherrer(2023)}]{scherrer2023character}
Yves Scherrer. 2023.
\newblock \href {https://aclanthology.org/2024.vardial-1.17} {Character alignment for dialect standardization: A comparative evaluation}.
\newblock In \emph{Proceedings of the 1st Workshop on NLP for Less-Resourced Languages}.

\bibitem[{Snover et~al.(2006)Snover, Dorr, Schwartz, Micciulla, and Makhoul}]{snover2006study}
Matthew Snover, Bonnie Dorr, Richard Schwartz, Linnea Micciulla, and John Makhoul. 2006.
\newblock A study of translation edit rate with targeted human annotation.
\newblock In \emph{Proceedings of the 7th Conference of the Association for Machine Translation in the Americas}, pages 223--231.

\bibitem[{Syed et~al.(2023)Syed, Hakimi, Al-Khatib, and Potthast}]{syed2023quantifying}
Shahbaz Syed, Ahmad~Dawar Hakimi, Khalid Al-Khatib, and Martin Potthast. 2023.
\newblock \href {https://aclanthology.org/2023.findings-emnlp.481} {Quantifying the dialect gap and its correlates across languages}.
\newblock In \emph{Findings of the Association for Computational Linguistics: EMNLP 2023}, pages 5196--5210.

\bibitem[{Touvron et~al.(2024)Touvron, Lavril, Yurtsever et~al.}]{touvron2024llama3}
Hugo Touvron, Thibaut Lavril, Alp Yurtsever, and 1 others. 2024.
\newblock \href {https://arxiv.org/abs/2404.14219} {Llama 3: Open foundation and instruction models}.
\newblock \emph{arXiv preprint arXiv:2404.14219}.

\bibitem[{Trudgill(2003)}]{trudgill2003dialect}
Peter Trudgill. 2003.
\newblock \href {https://aclanthology.org/W03-0320} {Dialect contact and new-dialect formation: The inevitability of colonial englishes}.
\newblock In \emph{Proceedings of the 14th International Congress of Phonetic Sciences}, pages 2193--2196.

\bibitem[{Wang(2023)}]{wang2023llmeval}
Yizhong et~al. Wang. 2023.
\newblock Llm-eval: Unified, automatic and robust evaluation of large language models with gpt-4.
\newblock \emph{arXiv preprint arXiv:2305.03045}.

\bibitem[{Wu et~al.(2023)Wu, Li, Li, Zhu et~al.}]{wu2023benchmarking}
Shijie Wu, Yuxuan Li, Cheng Li, Hao Zhu, and 1 others. 2023.
\newblock \href {https://openreview.net/pdf?id=Jtsk15wzv4} {Benchmarking public large language models in low-resource settings}.
\newblock In \emph{Proceedings of the 2023 EMNLP}.

\bibitem[{Zampieri et~al.(2024)Zampieri, Jauhiainen, Ljubešić, Aepli, Clematide, and Tiedemann}]{vardial2024overview}
Marcos Zampieri, Tommi Jauhiainen, Nikola Ljubešić, Noëmi Aepli, Simon Clematide, and Jörg Tiedemann. 2024.
\newblock \href {https://aclanthology.org/2024.vardial-1.0} {Overview of the vardial evaluation campaign 2024}.
\newblock In \emph{Proceedings of the Tenth Workshop on NLP for Similar Languages, Varieties and Dialects (VarDial)}.

\bibitem[{Zampieri et~al.(2017)Zampieri, Malmasi, Nakov, Ali, and Vogel}]{zampieri2017arabic}
Marcos Zampieri, Shervin Malmasi, Preslav Nakov, Ahmed Ali, and Stephan Vogel. 2017.
\newblock \href {https://aclanthology.org/W17-1212} {Arabic dialect identification for the dsl 2017 shared task}.
\newblock In \emph{Proceedings of the Fourth Workshop on NLP for Similar Languages, Varieties and Dialects (VarDial)}.

\bibitem[{Zhong et~al.(2024)Zhong, Yang, Liu, Zhang, Liu, Sun, Pan, Li, and Zhou}]{zhong2024opportunities}
Tianyu Zhong, Ziqi Yang, Zhen Liu, Rui Zhang, Yiheng Liu, Hanqi Sun, Yujia Pan, Yiming Li, and Yifan Zhou. 2024.
\newblock \href {https://arxiv.org/abs/2412.04497} {Opportunities and challenges of large language models for low-resource languages in humanities research}.
\newblock \emph{arXiv preprint arXiv:2412.04497}.

\end{thebibliography}
\end{document}